# PTGCF: Printing Texture Guided Color Fusion for Impressionism Oil Painting Style Rendering

**Jing Geng[1#*], Li'e Ma [1#*], Xiaoquan Li[2], Yijun Yan[2*]**

**Abstract** As a major branch of Non-Photorealistic Rendering (NPR), image stylization mainly uses the computer algorithms to render a photo into an artistic painting. Recent work has shown that the extraction of style information such as stroke texture and color of the target style image is the key to image stylization. Given its stroke texture and color characteristics, a new stroke rendering method is proposed, which fully considers the tonal characteristics and the representative color of the original oil painting, in order to fit the tone of the original oil painting image into the stylized image and make it close to the artist's creative effect. The experiments have validated the efficacy of the proposed model. This method would be more suitable for the works of pointillism painters with a relatively uniform sense of direction, especially for natural scenes. When the original painting brush strokes have a clearer sense of direction, using this method to simulate brushwork texture features can be less satisfactory.

## I. Introduction

Non-Photorealistic Rendering (NPR) is one of the important branches of computer graphics. Unlike realistic painting, which pursues physical accuracy and painting authenticity, non-photorealistic painting (NPP) draws scenes through artistic expression techniques, such as oil painting, pencil drawing, cartoon et al, so that the painting effect can convey rich feelings [1]. Among various NPP styles, oil painting is widely favored for its long history and rich expressiveness. It can not only accurately depict the scene, as shown in Fig. 1(a), but also convey various emotions of the painter through different artistic exaggerations (Fig. 1 (b)).

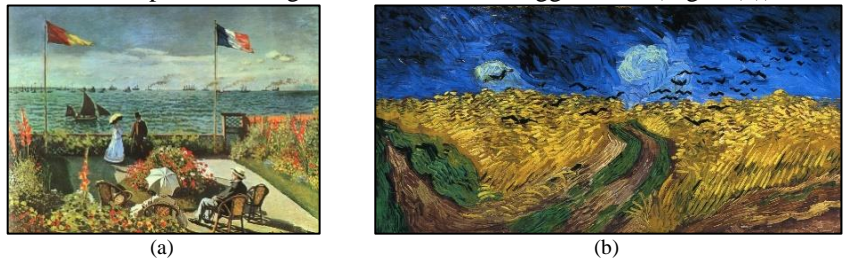

(a) (b)
Fig. 1. (a) Monet, "Terrace at St. Adresse, 1866", (b) Gogh, "Wheatfield with Crows 1890".

In Fig.2, an example is given, where a photo is converted by a style automatically extracted from an oil painting by a professional artist: Monet.

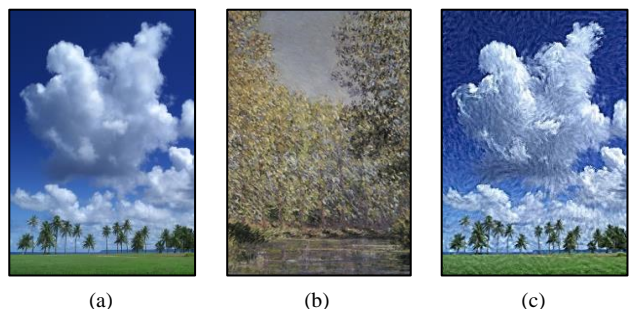

(a) (b) (c)
Fig. 2. Sample of image rendering: (a) Landscape, (b) Monet, "Bend in the Epte River near Giverny", (c) Stylized image.

✉ J. Geng, Ma Li'e, Y. Yan
e-mail: gengjing_qianye@163.com, malie@xaut.edu.cn, ggaa900301@gmail.com
[1]Faculty of Printing, Packaging Engineering and Digital Media Technology, Xi'an University of Technology, Xi'an, China
[2] Department of Electronic and Electrical Engineering, University of Strathclyde, Glasgow, UK
[#]Jing Geng and Ma Li'e equally contributed to this work

In the history of art development, Impressionism is a very important categories of painting. This category pays attention to the true description of nature and the light and shadow changes of things, which has played a good transitional role in the transformation of European painting from realism to modernism. It is precisely because of these unique artistic pursuits that the works of this category have different aesthetic characteristics against any other period and painting categories in terms of subject matter, color, composition and other aspects.

Therefore, the protection of these paintings has become a problem to be addressed urgently. However, how to digitize these antique priceless paintings and exhibit them on the Internet is reported as a challenging problem due to the following reasons.

There are many ways to realize image stylization. According to the timeline, it is divided into two categories. Traditional image style migration, image style transfer based on Deep learning network. Due to the characteristics of Impressionism art and similar techniques presented in many Impressionism artists' artworks, it is usually difficult for researchers to fully extract the distinctive features from each painter's work. With deep learning, this drawback seems can be somehow overcome, yet at a cost of a large amount of data requested for training to learn the prior knowledge [2]. To address these problems, we first develop an adaptive brush strokes selection approach to automatically get a generic texture path of an oil painting image. Then, the color and texture feature will be integrated in the Fourier domain. Eventually, the stylized image will be generated by inverse Fourier transform. A comprehensive experiment with detail analysis will be also carried out.

Overall, the main contributions of this paper can be summarized as follows:
1. We proposed an unsupervised texture guided color fusion framework, namely TGCF, for impressionist oil painting style migration, which fully consider the tonal characteristics and the representative color of the original oil painting.
2. We proposed an effective fusion strategy to adaptively transfer the texture information from oil painting image to the nature image in the frequency domain of dominant color component.

The remaining of our paper is organized as follows. In Section II, we give a brief review of the related work. In Section III, the architecture of the proposed TGCF is elaborated in detail. In Section IV, we perform abundant experiments and analyzed the effects of oil painting style images with different themes. Finally, some concluding remarks are drawn in Section V.

## II. RELATED WORK

Image style transfer is an image processing method that renders the semantic content of an image with different styles. In the field of NPR, image art style rendering technology can be divided into three categories: methods based on stroke rendering, methods based on image analogy and methods based on image filtering. Efros [3] synthesized the target image by extracting and reorganizing texture samples; Hertzman [4] transformed the existing image style into the target image through image analogy; Ashikhmin [5] transformed the high-frequency texture of the source image into the target image while retaining the coarse scale of the target image; Lee [6] improved Ashikhmin's algorithm by passing additional edge information.

In recent years, with the wide application of deep neural network in other visual perception fields (such as object and face recognition), it also has good application in the field of style transfer. Image style transfer methods of deep learning mainly include image-based iteration and model-based iteration [7]. Image iteration is an optimization iteration directly on white noise image to realize style transfer, and its optimization goal is image; the optimization goal of model iteration is the neural network model, which realizes style transfer in the way of network feedforward.

There are three main types of image migration algorithms based on Markov mean and iterative style. It has the characteristics of high quality of synthetic image, good controllability, easy parameter adjustment, no training network, long calculation time and great dependence on pre training model.

Gatys [8] first applied VGG19 network to style migration in 2015. The key discovery of Gatys is that the content and style of convolutional neural network are separated and the style feature representation of any image can be extracted by constructing Gram matrix. Because the color in the migration style will destroy the original image content information, Gatys proposed two style migration methods to save the color information of the original image [9]: color histogram matching (RGB color space) and brightness migration only (L*a*b* color space).

Gatys used the filter pyramid in VGG network as the high-level representation of the image. However, different convolution layers can only capture the connection between each pixel of the image, but lack the connection in spatial distribution. Based on the above problems, Li et al. proposed a model combining Markov random field and convolutional neural network [10], replacing the Gram matrix matching in Gatys model with Markov regularization model. Markov model describes the set of similar feature information, so the CNN image feature map is divided into many regional blocks and matched, which can improve the visual rationality of the synthetic image. This method provides an interesting extension for style transfer based on image iteration, but it also has many limitations: (1) There are strong restrictions on the input image: the style image must be able to be reconstructed by Markov. For example, a picture with a strong transmission structure is not suitable for this method. (2) As a result, compared with the original image, the image will have the disadvantage of fuzzy edge. This is because of the loss of images in the process of network training. To sum up, the style transfer method based on Markov random field will produce good results only when the content image and style image have similar shapes without strong angle and size changes.

Liao [11] achieved good results by combining deep learning (VGG19) and image analogy (patchmatch) for style transfer. Liao et al. first pre trained a VGG19 network to extract image features through deep learning, and then extended patchmatch from image domain to feature domain, effectively guiding the semantic level visual attribute migration. The algorithm lacks the ability of semantic generalization, but it can only transfer similar content.

Although the method based on deep learning can produce excellent style synthetic images, it has the problem of low efficiency. The image style migration method based on model iteration generates the model through a large amount of data training. The style migration algorithm based on model iteration mainly includes style migration based on generation model and style migration based on image reconstruction decoder.

Johnson [12] first proposed the image style migration method of iterative optimization generation model, namely fast style migration. Compared with the loss function compared pixel by pixel in the previous training to generate the model, the perceptual loss function squares the high-level abstract feature representation extracted by the pre training VGG model. The solution of this part of the problem is consistent with the algorithm of Gatys et al.

In general, deep learning based rendering methods have two disadvantages: parameter adjustment and low efficiency. Although fast style transfer alleviates the problem of low efficiency, it can only carry out model training for specific styles, and still can't avoid the problem of parameter adjustment.

Therefore, according to the characteristics of Impressionist oil painting, we introduce a generic image style migration method, based on the rendering of physical model and texture synthesis, and propose an efficient NPR framework for adaptively fusing the texture and color attributes of the target oil painting image to realize the image stylization of impressionist style. Firstly, an adaptive stroke selection method is introduced to automatically obtain the general texture block of oil painting stylized image. Then, the color and texture information of the source image and stroke are fused in the Fourier domain. By making full use of the color and texture features of oil painting stylized images, the source images can be imitated into target images with any artist's style, such as Monet, Shura and Van Gogh. In order to select the best parameters to obtain the best rendering performance, the selection of pen block size is discussed, and the influence of different painters' attributes on stylized images is analyzed.

### III. MATERIAL AND METHODS

#### A. Data collection and pre-processing

The experimental image data are acquired by a D65 Kodak camera with 12 mega pixels CMOS sensor. The original digital photos mainly cover the real landscape and portrait in our life. The examples of acquired data are shown in Fig. 3. Due to the inconsistent lighting issue in different scenes, the tone reproduction of each captured images was different, which may cause inconsistent image stylized effects. To solve this problem, a simple yet fast image pre-processing algorithm [13] was used to adjust the chroma and contrast of digital images.

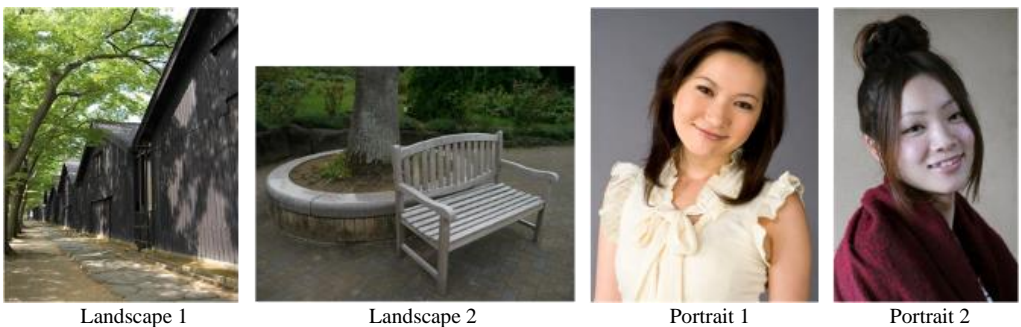

| Landscape 1 | Landscape 2 | Portrait 1 | Portrait 2 |

Fig. 3. The original digital photos for stylized migration

#### B. Workflow of the proposed TGCF model

Fig. 4 shows the workflow of the proposed method, which is composed of three data processing modules, i.e. color space conversion, color matching, and feature fusion. Color space conversion is a standard procedure of coordinate transformation between RGB and L*a*b* color space. It is an essential step in the color matching and also plays a vital role between color matching and feature fusion. Color matching aims to enhance the original digital images and make its appearance similar to the impressionism oil painting. In feature fusion, the color attributes of enhanced image and brush stroke will be adaptively fused in the L*a*b* color space according to their texture attributes. The final stylized RGB image will be generated after color space conversion. The implementation of color matching and feature fusion will be detailed below.

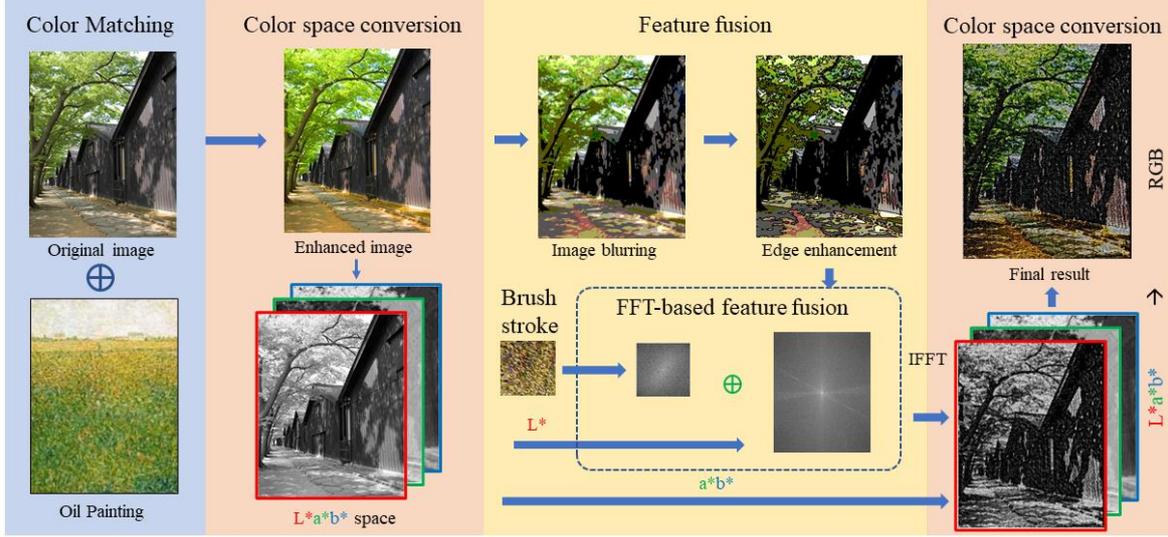

Fig. 4. Process of producing oil painting-style images made by non-photorealistic rendering

*C. Color Matching*

To introduce the impressionism style into the original digital photos, the first step is to imitate the color attributes of the oil painting at the original digital photos. For this purpose, a color transfer between images algorithm [14] is employed. The implementation of color matching [19] is detailed below:

1) Input a source and an oil painting image. In the Fig.5, the landscape image on the left is the source image, the middle image is the oil painting image;

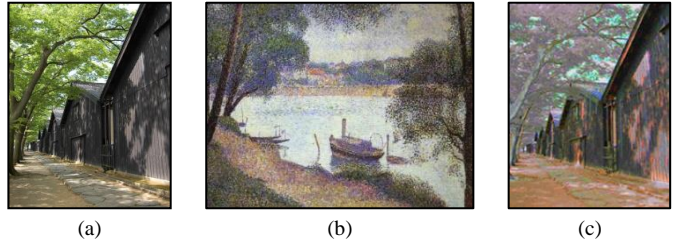

(a)             (b)             (c)
Fig. 5. Image samples, (a) source image, (b) oil painting image, (c) image after color conversion.

2) Convert both the source and the oil painting image from the RGB to the Lab color space;
3) Calculate the mean value and standard deviation of each channels for the source and oil painting images in Lab color space, which are denoted as $(\mu_{l.s}, \mu_{a.s}, \mu_{b.s}, \sigma_{l.s}, \sigma_{a.s}, \sigma_{b.s})$ and $(\mu_{l.o}, \mu_{a.o}, \mu_{b.o}, \sigma_{l.o}, \sigma_{a.o}, \sigma_{b.o})$;
4) Subtract the mean value of each channel (i.e., $l_o$, $a_o$, $b_o$) from the oil painting image in Lab color space;

$$l' = l_o - \mu_{l.o}, \; a' = a_o - \mu_{a.o}, \; b' = b_o - \mu_{b.o} \tag{1}$$

5) Scale the oil painting channels by the ratio of the standard deviation of the oil painting image divided by the standard deviation of the source image, multiplied by the channels of oil painting image;

$$l_{scale} = \frac{\sigma_{l.o}}{\sigma_{l.s}} l', \; a_{scale} = \frac{\sigma_{a.o}}{\sigma_{a.s}} a', \; b_{scale} = \frac{\sigma_{b.o}}{\sigma_{b.s}} b' \tag{2}$$

6) Add the mean value of the result of previous step into the source image;

$$l_{new} = \overline{l_{scale}} + l_s, \; a_{new} = \overline{a_{scale}} + a_s, \; b_{new} = \overline{b_{scale}} + b_s \tag{3}$$

7) Convert the result of previous step from the L*a*b* space to the RGB color space.Feature Fusion

After color matching, the texture of enhanced digital images and oil painting images will be analysed, which will be used to further guide the color fusion procedure for final optimized image rendering.

*1) Edge enhancement*

For each enhance digital image, the edge attributes will be extracted in the RGB color space. Then USM algorithm [15] is employed to enhance and extract the edge.

Implementation steps:

Step 1: Read in the pixel data $D_1$ of the source image.

Step 2: Obtain the smoothed pixel data $D_2$ using Gaussian blur.

Step 3: According to the input parameter w, for each pixel on the image, use the USM sharpening formula to calculate the pixel after sharpening each pixel. The USM sharpening formula is expressed as follows:

$$(D_1 - w * D_2) / (1 - w) \qquad (4)$$

where $w \in [0.1, 0.9]$ represents the weight and we set it as 0.6 for fast implementation in this paper.

Step 4: Return the enhanced image.

*2) Adaptive selection of brush stroke*

Brush strokes refer to the markers produced by the painter in the process of painting, which can be used to describe the objective object or to express the subjective emotion. In this sub-section, the brush strokes information will be extracted from an oil painting image. For each oil painting image, the brush stroke texture patch is automatically decided by calculating the standard deviation ($SD(i)$) in the sliding windows ($W_{x,y}(i)$), where *x* and *y* are the spatial size of the window.

$$SD(i) = \sqrt{\frac{1}{N}\sum_{j=1}^{N}\left(o_j(i) - \mu(i)\right)^2} \qquad (5)$$

Here we have $i \in [1, I]$, $I$ is the number of sliding windows with the stride of 64 pixels. $N = x * y$, $o_j(i)$ represents the $j^{th}$ pixels in the $i^{th}$ sliding window, $\mu(i)$ denotes the mean pixel value in the $i^{th}$ sliding window.

The sliding window with the lowest SD will be selected as brush stroke texture patch. 65 original paintings were studied, including 10 Van Gogh's landscapes, 12 Van Gogh's portraits, 15 Monet's landscapes, 12 Monet's portraits, 9 Seurat's landscapes, and 7 Seurat's portraits. The method has a certain universality. Some representative artworks of each artist are shown in Fig.6, and the rest artworks are shown in the appendix.

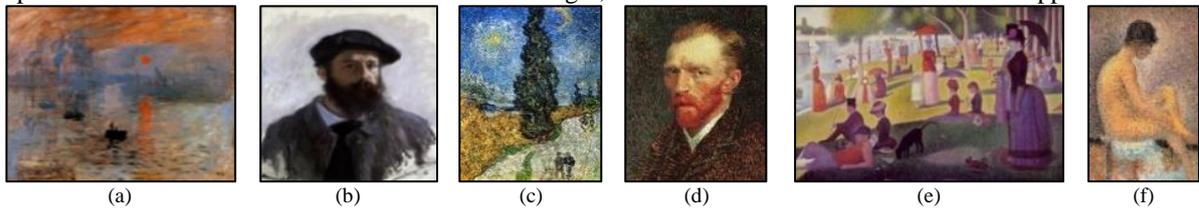

(a) (b) (c) (d) (e) (f)

Fig. 6. (a) Monet, "Impression Sunrise", (b) Monet, "Self-Portrait", (c) Van Gogh, "Cypress against a Starry Sky", (d) Van Gogh, "Self-Portrait", (e)Seurat, "A Sunday Afternoon on the Island of La Grande Jatte", (f) Seurat, "Model".

In order to gather our collection of fine-art paintings, we used the publicly available dataset of "Wikiart paintings"[1]; which, to the best of our knowledge, is the largest online public collection of digitized artworks. Examples of selected brush strokes in this study are shown in Fig. 7, where Fig. 7 (a), (c) and (e) show the oil painting of "Bend in the Epte River near Giverny" by Claude Monet, "Cypress against a Starry Sky" by Van Gogh and "Models.Detail" by Seurat, respectively. Fig. 7 (b), (d) and (f) show the selected brush strokes of the three oil paintings.

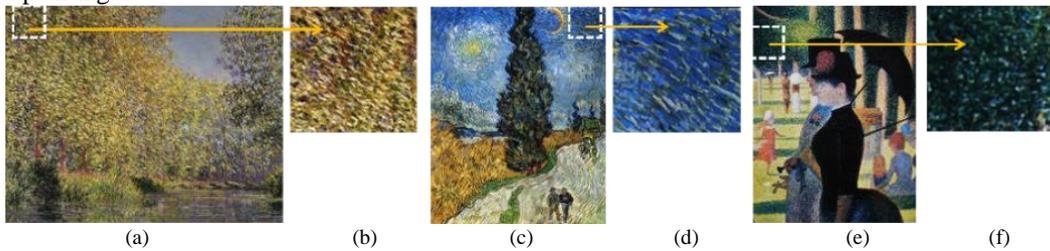

(a) (b) (c) (d) (e) (f)

Fig. 7. (a) Monet, "Bend in the Epte River near Giverny", (b)Brush stroke image, (c)Van Gogh, "Cypress against a Starry Sky", (d)Brush stroke image, (e)Seurat, "Models.Detail", (f) Brush stroke image.

---

[1] http://www.wikiart.org/

*3) Feature fusion in FFT domain*

For an actual oil painting artwork, the brush stroke direction and the gradient information are the most important texture attributes, respectively. As shown in Fig. 8, for a single orange, the brush stroke direction and gradient of its surface and surrounding shadows have already shown the uniqueness. Therefore, a piece of oil painting will contain more abstract stroke patterns. Unlike oil painting image, digital image doesn't have brush stroke direction, but the gradient information is still vital to characterise the texture in the image.

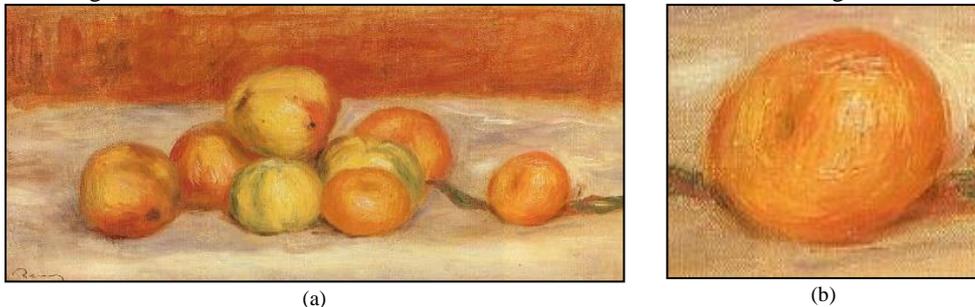

(a)     (b)

Fig. 8. (a) "Apples and Manderines" Renoir(impressionism), (b) enlargement of selected region.

Therefore, we take the full usage of texture information in the selected brush as well as the digital image, then guide the adaptive color information fusion of brush stroke and digital image in order to fully render the oil painting style on the digital image. The detailed implementation is presented below.

Given a enhance digital image and a selected brush stroke, they will be converted from RGB color space to L*a*b* color space. Then, their L* components will be transferred to the frequency domain using the fast Fourier transformation (FFT).

As seen in Fig. 9, a power spectrum for each phase angle can be built after transferring the brush stroke into FFT domain. Then the angle of maximum power can be extracted.

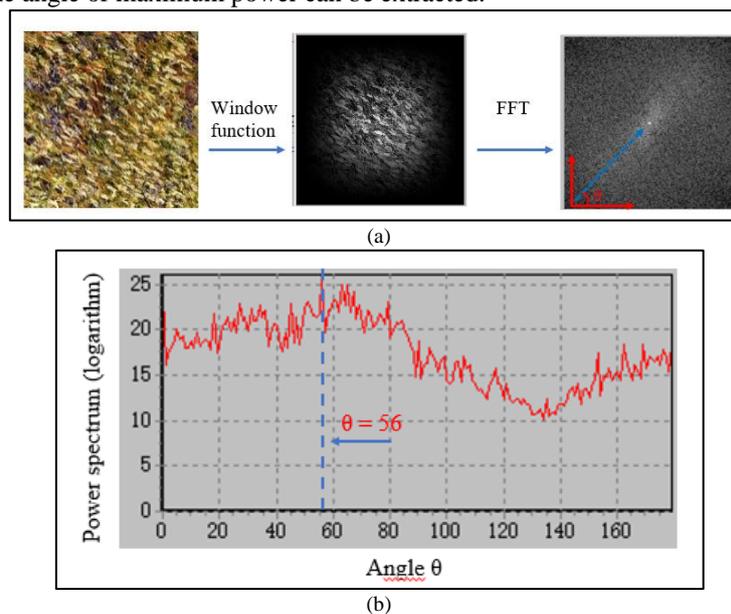

(a)

(b)

Fig. 9. (a) Calculate the average value of the power spectrum of each angle, (b) Monet stroke direction information

On the other hand, the gradient information of each digital image can be extracted after edge enhancement. For each pixel window in the FFT domain, its value will be updated after shifting the brush stroke with the specific phase angle (Fig. 10).

Based on this fact, we came up with an algorithm to add stroke texture to digital photos to produce oil painting style images. Algorithm adapted to stroke (gradient perpendicular to photo gradient)

   The summary is as follows.
1) Convert the original photograph into a frequency domain.
2) Convert the brush texture patch into a frequency domain to create a filter.
3) Blur the original photo.
4) Enhance the boundary of image of step 3.
5) Acquire the direction information of gradation from the output of step 4.
6) Update the value of original photograph by adaptively shifting the brush stroke in FFT domain as shown in

Fig. 10 (b).
7) Convert the result of step 6 into spatial domain by inverse Fourier transformed.
8) Concatenate the result of step 7 with original a * b *. To this end, the color and texture features of digital image and brush stroke have been successfully fused.
9) Finally, the fused result of step 8 will be transferred from L*a*b* to the RGB color space for displaying the rendering effects.

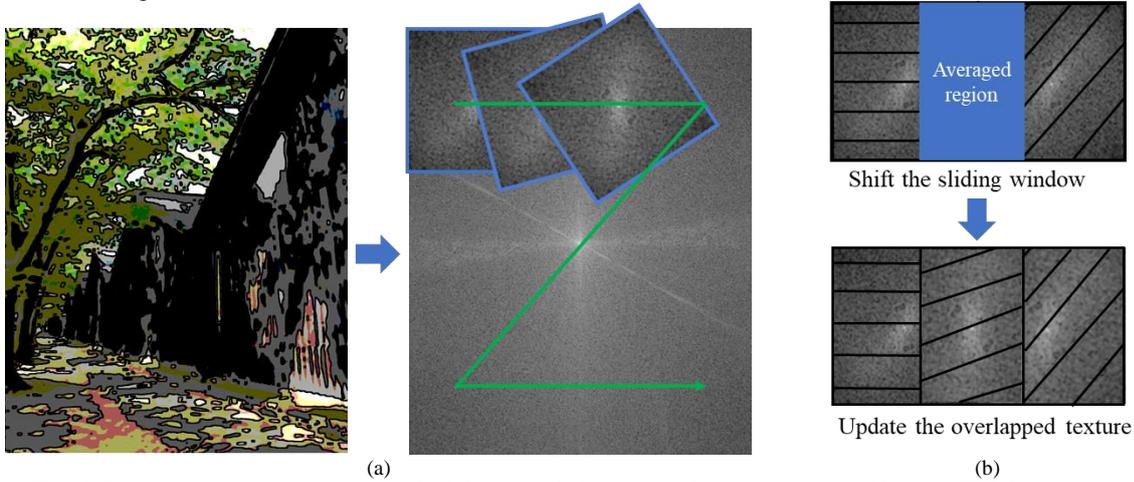

(a) (b)
Fig. 10. Frequency domain texture guided color information fusion, (a) transfer the edge enhanced image to FFT domain, (b) shift the sliding window and update the value in FFT domain based on phase angle.

### D. Discussion of stroke texture

From the movement of stroke texture, it is naturally associated with a problem. That is, how to shift the sliding window to fuse stroke texture with enhanced image in order to make the generated image closer to the oil painting. To choose the best size of sliding window, we have compared the performance using three moving strategies, i.e., $\frac{1}{8}$, $\frac{1}{4}$ and $\frac{1}{2}$ of the patch size in the horizontal and vertical directions respectively (Fig. 11).

The evaluation image is printed on photo (glossy) paper with light black ink in the inkjet printer Epson px-5500. The printing size is 12cm*16cm. Under the high color rendering fluorescent lamp of D50, the brightness of 700lx is evaluated. There is no limit to the observation distance. The subjective evaluate was done by 30 students aged between 20 and 30. The participants were asked to compare the stylized image with Monet's collection of paintings.

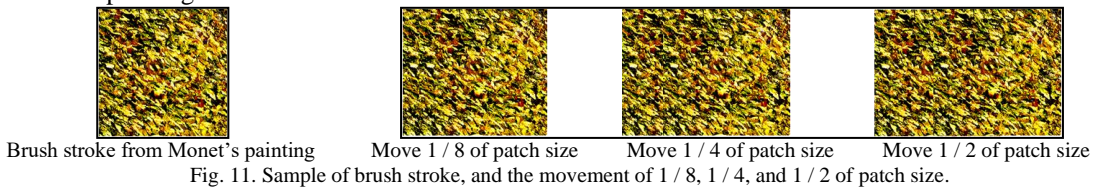

Brush stroke from Monet's painting    Move 1 / 8 of patch size    Move 1 / 4 of patch size    Move 1 / 2 of patch size
Fig. 11. Sample of brush stroke, and the movement of 1 / 8, 1 / 4, and 1 / 2 of patch size.

The evaluation method is Rank order method [16]. This method is to present many stimuli at the same time,

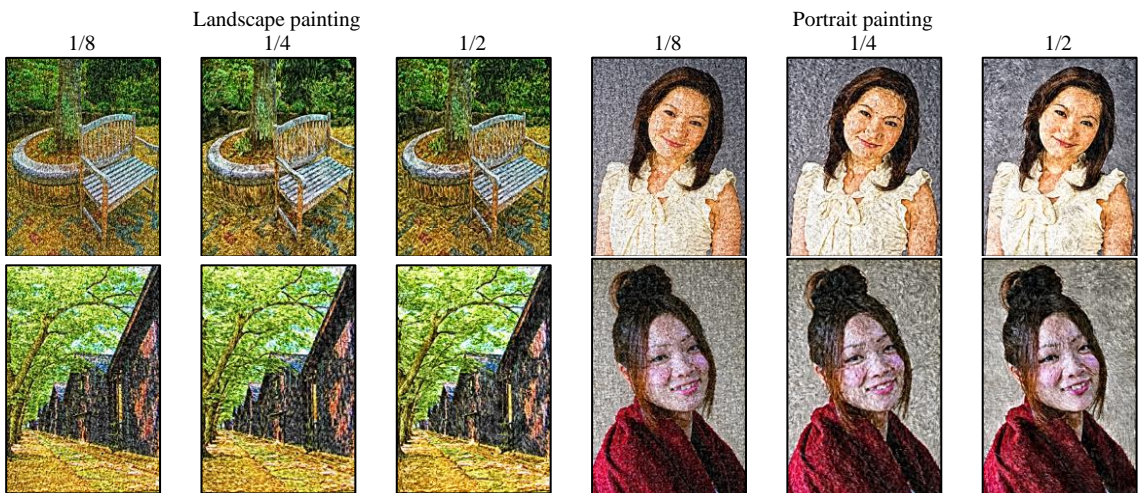

Fig. 12. generated examples of landscape and portrait painting with the movement of 1 / 8, 1 / 4, and 1 / 2.

where many participants will rank these stimuli in order according to certain standards, and then the averaged rank score on each stimulus can be obtained. In our study, participants were asked to rank the three oil painting style images according to the criteria of artist-style-similarity between the rendered images and original oil paintings. After massive subjective evaluation, the average rank score of rendered images can be obtained, which can reflect the rendering performance. The stroke texture block to be applied moves 1 / 8, 1 / 4 and 1 / 2 of the patch size horizontally and vertically, and the three generated images are shown in Fig. 12. The scale value representing the degree of compliance with the evaluation benchmark of each stimulus is obtained. For landscape 1 and portrait 1, the selection of 1 / 4 and 1 / 2 for the patch size gives the same score. On the other hand, for landscape 2 and portrait 2, the selection of 1 / 8 for the patch size is not ideal. In this way, if the number of overlapping becomes less, the number of mixing colors becomes less.

## IV. RESULTS AND DISCUSSION

### A. Landscape painting

"Starry Night" is one of the most well-known paintings by Van Gogh (Fig.13). There are three dominant color: sky blue, yellow, and black. In general, the appearance of stylized image is close to the original oil painting images. However, some elements of the oil painting image (i.e., orange moon and the cluster of starligh) don't reflect on the rendering effects. This is mainly because this painting consists of various texture such as curved long lines, dash lines, swirl patterns, etc. However, the brush stroke texture patch doesn't include such information, which indicates that the selected brush stroke can merely imitate the general tone and ignore the less-dominant colours.

Impressionism believes that reality is a fleeting visual impression, which breaks through the previous painting stereotypes which is the usage of high contrast color. In Haystacks series (Fig.14(b)), Monet used technique of juxtaposing solid and dot colors, which can reflect the light change in the scene. Although the image stylization result (Fig.14(c)) can imitate a brown tone of the oil painting, the effect of light change and shadow is insufficient. This can be possibly mitigated by mathematical modelling of computer graphics.

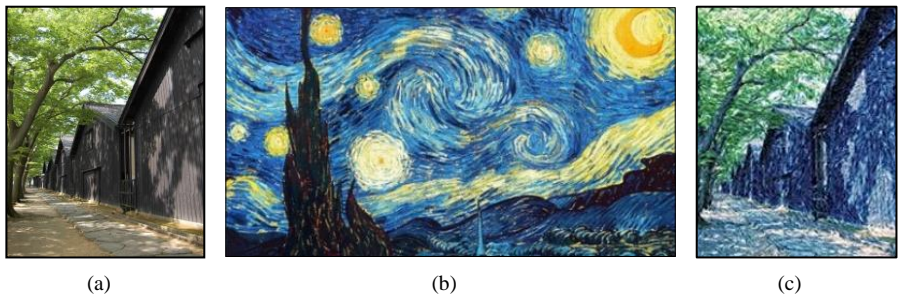

(a)                (b)              (c)

Fig. 13. (a) Landscape, (b) Van Gogh, "Starry Night", (c) Stylized image.

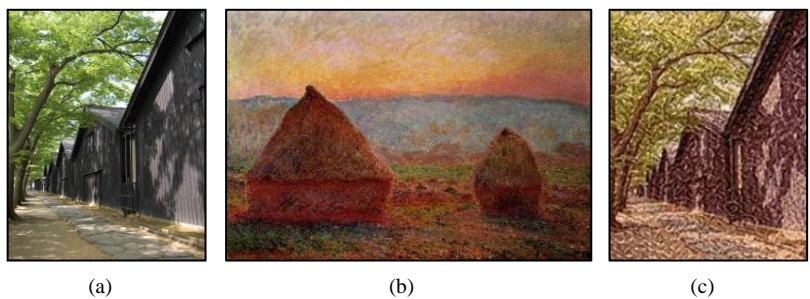

(a)                (b)              (c)

Fig. 14. (a) Landscape, (b) Monet, " Haystacks series ", (c) Stylized image.

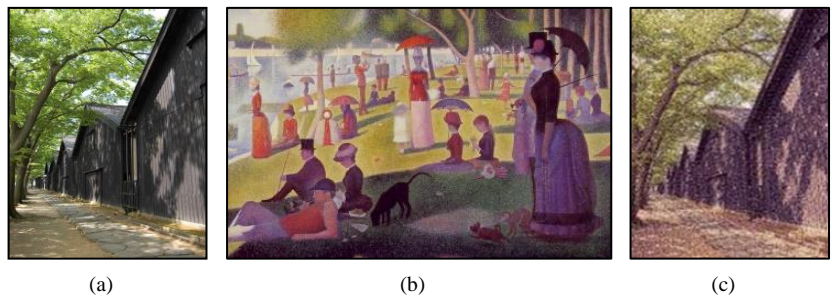

(a)                (b)              (c)

Fig. 15. (a) Landscape, (b) Seurat, "Big Bowl Island Sunday Afternoon", (c) Stylized image.

At the end of the 19th century, French painter Georges Seurat was the first one to propose and practice the stippling and color separation painting method which uses a unique stroke to generate colors by primary colors. "Big Bowl Island Sunday Afternoon" is a representative work of Neo-Impressionism (Fig. 15(b)), where the purple color is mixed by parallelly painting the red and blur colors. In addition, this oil painting is composed of millions of widely distributed color dots, which brings the difficulty of fully imitating its color and texture information.

### B. Portrait painting

The similarity between the final rendering effect of oil painting stylization and the oil painting image is high. However, due to the lack of semantics of the image content, the rendering effects on the portrait are far away from our human perception. As shown in the third row of Fig. 16. Van Gogh's paintings are formed by thick and powerful strokes, intense, heavy, strong and distorted color lines. However, when Van Gogh's powerful stroke texture is applied to the portrait, the performance of the skin, the sticky touch of the clothes and the overall texture of the image are affected, which significantly reduce the visualisation. To tackle with this problem, abstract and high-level sematic features can be extremely useful, which can be achieved by multiple convolutional layers with large perception field and pyramid pooling.

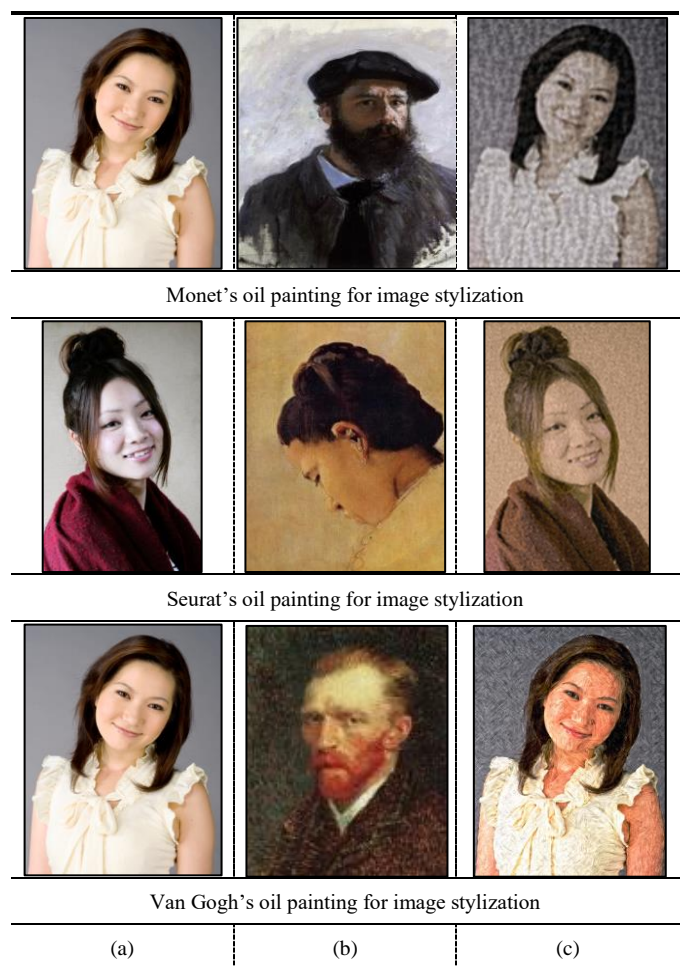

Monet's oil painting for image stylization

Seurat's oil painting for image stylization

Van Gogh's oil painting for image stylization

(a)     (b)     (c)

Fig. 16. Sample of image stylization: (a) portrait image, (b) artists' oil painting, (c) stylized image.

## V. CONCLUSIONS

In this paper, an effective model TGCF is proposed for impressionism-focused image stylization, which takes full use of both the representation color and the brush stroke texture of the oil paintings and produce much improved the visual effect of the produced results. Four source photographs were selected: two landscapes and two portraits, and 65 oil paintings were studied, including 10 Van Gogh's landscapes, 12 Van Gogh's portraits,15 Monet's landscapes, 12 Monet's portraits, 9 Seurat's landscapes, and 7 of Seurat's portraits. This has shown its generalization in dealing with various artist styles. For our proposed model, it is easy to learn some Impressionist features, which are difficult to learn for other algorithm models, or other models focus more on universal oil painting. This may be because the trajectory of texture rendering used in the model TGCF is based on Impressionism.

According to the current research trends, there are still further improvements that can be made in the process of image stylization. How to judge whether the rendered image belongs to the expected style. Deep learning based analysis can be applied, such as DenseNet [17] model and MTFFNet [18] model. Therefore, our next work is to establish an improved deep learning network to objectively evaluate the classification of rendered images, achieve the approximation between rendered images and real oil paintings.

Another work would focus on semantic segmentation based image stylization. Semantic information about image content is seldom considered in image stylization. Most of the methods extract the low-level formal features of the image (for example, the direction field of the segmentation block determines the position of the strokes) while little consideration is given to the content information of the image, which may lead to an inability to achieve the desired effect regarding the position and direction of the brush strokes.

Moreover, there is a lack of distinction between the important and non-important elements of the image. Stylization, such as various forms of filtering operations, abstracts graphics to a certain extent. Therefore, content-aware image stylization can be a new insight in the future.

**Funding** This research work was financially supported by the Application of multi-sensor image information fusion technology in fountain landscape image, National Key Research and Development Program of China, Technology Innovation Leading Program of Shaanxi Province. Project Number:108/441219001, 2019YFB1707200, 2020QFY03-04.

The images of "Portrait1 and Portrait2" in the manuscript are photos of former colleagues of the first author in Japan, where consent was received to publish the image in the research paper.

**Compliance with Ethical Standards**

**Conflict of Interest:** The authors declare no conflict of interest.

REFERENCES

[1] A. Lake, C. Marshall, M. Harris, and M. Blackstein, "Stylized rendering techniques for scalable real-time 3d animation," Proceedings of the 1st international symposium on Non-photorealistic animation and rendering, pp. 13-20, 2000.
[2] Y. Yan et al., "Unsupervised image saliency detection with Gestalt-laws guided optimization and visual attention based refinement," Pattern Recognition, vol. 79, pp. 65-78, 2018.
[3] A.Efros , and W. Freeman . "Image quilting for texture synthesis," Proceedings of the 28th annual conference on Computer graphics and interactive techniques, pp. 341-346, 2001.
[4] Jh Jari .Huttunen, "Image Analogies." Proceedings of Acm Siggraph Acm Press 293(2004). pp. 327-340.Huttunen, Jh Jari . "Image Analogies." Proceedings of Acm Siggraph Acm Press 293, pp. 327-340, 2004.
[5] Ashikhmin, and Michael. "Fast Texture Transfer, " IEEE Computer Graphics & Applications, vol. 23(4), pp. 38-43, 2003.
[6] H. Lee, et al. "Directional texture transfer." International Symposium on Non-photorealistic Animation & Rendering ACM, pp. 43-48, 2010.
[7] S. Chen, et al. "Survey of image style transfer based on deep learning," Application Research of Computers pp. 2250-2255, 2019.
[8] L. A. Gatys,  A. S. Ecker , and  M. Bethge . "A Neural Algorithm of Artistic Style," Journal of Vision, vol. 16, pp. 326, 2016.
[9] L. A. Gatys, et al. "Preserving Color in Neural Artistic Style Transfer," arXiv preprint arXiv:1606.05897, 2016.
[10] C. Li, and  M. Wand . "Combining Markov Random Fields and Convolutional Neural Networks for Image Synthesis," IEEE Conference on Computer Vision and Pattern Recognition, pp. 2479-2486, 2016.
[11] J. Liao, et al. "Visual attribute transfer through deep image analogy," ACM Transactions on Graphics, vol. 36(4), pp.1-15, 2017.
[12] J. Johnson , A. Alahi , and  F. Li, "Perceptual Losses for Real-Time Style Transfer and Super-Resolution," European conference on computer vision pp. 694-711, 2016.
[13] E. Reinhard, M. Stark, P. Shirley, and J. Ferwerda, "Photographic tone reproduction for digital images," pp. 267-276, 2002.
[14] E. Reinhard, M. Adhikhmin, B. Gooch, and P. Shirley, "Color transfer between images," IEEE Computer graphics and applications, vol. 21, no. 5, pp. 34-41, 2001.
[15] A. Polesel, et al. "Image Enhancement via Adaptive Unsharp Masking," IEEE Transactions on Image Processing, vol. 9(3), pp. 505-510, 2000.
[16] P. G. Engeldrum, Psychometric scaling: a toolkit for imaging systems development. Imcotek press, 2000.
[17] G. Huang, Z. Liu, L. Van Der Maaten, and K. Q. Weinberger, "Densely connected convolutional networks," Proceedings of the IEEE conference on computer vision and pattern recognition, pp. 4700-4708, 2017.
[18] W. Jiang et al., "MTFFNet: a Multi-task Feature Fusion Framework for Chinese Painting Classification," Cognitive Computation, vol. 13, no. 5, pp. 1287-1296, 2021.
[19] Super fast color transfer between images, Access at https://pyimagesearch.com/2014/06/30/super-fast-color-transfer-images/, 2014